\documentclass[pre,amsmath,amssymb,aps,floatfix,nofootinbib]{revtex4-2}

\usepackage{amsmath}
\usepackage{graphicx}
\usepackage{epstopdf}%
\usepackage{multirow}
\usepackage{tabularx}

\usepackage{hyperref} 
\usepackage{url} 

\usepackage{booktabs}
\usepackage{tabularx}
\usepackage{appendix}
\usepackage{xcolor}

\newcommand{\comment}[1]{}

\newcommand{\BEQ}{\begin{equation}}
\newcommand{\EEQ}{\end{equation}}
\newcommand{\BEA}{\begin{eqnarray}}
\newcommand{\EEA}{\end{eqnarray}}

\newcommand{\1}{$^{(1)}$}
\newcommand{\2}{$^{(2)}$}
\newcommand{\3}{$^{(3)}$}
\newcommand{\4}{$^{(4)}$}
\newcommand{\5}{$^{(5)}$}
\newcommand{\6}{$^{(6)}$}
\newcommand{\7}{$^{(7)}$}
\newcommand{\8}{$^{(8)}$}
\newcommand{\9}{$^{(9)}$}
\newcommand{\0}{$^{(10)}$}

\newcommand{\perm}{{\rm perm}}

\newcommand{\kk}{{\rm k}}
\newcommand{\nk}{{\rm nk}}

\begin{document}

\title{Unsupervised extraction of local and global keywords from a single text}

\author{Lida Aleksanyan and Armen Allahverdyan}
\affiliation{Alikhanyan National Laboratory, Alikhanian Brothers Street 2, Yerevan 0036, Armenia,\\
\email{armen.allahverdyan@gmail.com}}

\begin{abstract}
We propose an unsupervised, corpus-independent method to extract keywords from a single text. It is based on the spatial distribution of words and the response of this distribution to a random permutation of words. Our method has three advantages over existing unsupervised methods (such as YAKE, for example). First, it is significantly more effective at extracting keywords from long texts in terms of precision and recall. Second, it allows inference of two types of keywords: local and global. Third, it extracts basic topics from texts. 
Additionally, our method is language-independent and applies to short texts. The results are obtained via human annotators with previous knowledge of texts from our database of classical literary works. The agreement between annotators is moderate to substantial. Our results are supported via human-independent arguments based on the average length of extracted content words and on the average number of nouns in extracted words. We discuss relations of keywords with higher-order textual features and reveal a connection between keywords and chapter divisions. 
\end{abstract}

\maketitle

\section{Introduction}

Keyword extraction from texts is important for information retrieval and NLP tasks (document searching within a larger database, document indexing, feature extraction, and automatic summarization) \citep{manning,review4, review5}. As an analytical tool, keywords reflect the meaning of a text and help to extract its topics. Hence, keywords and their extraction schemes are also employed in discourse analysis \citep{keyness}. Our focus will be on this analytical aspect of keywords. 

Keyword extraction is challenging, as the state-of-the-art results demonstrate \citep{review1,review2,review3}. A possible explanation is the lack of a sufficiently comprehensive definition of the keyword concept. Keywords are generally non-polysemic nouns (i.e., nouns that do not have many sufficiently different meanings) related to text topics \citep{keyness}. 

Computer science approaches to keyword extraction fall into three main groups; see section \ref{related}. First, there are supervised methods, which usually require a large training set to learn which keywords should be found and where to find them \citep{gollapalli,genex,kpspotter,bert}. The second group involves unsupervised methods that demand a corpus of texts for contrastive learning \citep{Ando2005,tfidf1,tfidf2,textual}. The third group involves methods that apply to a single text, i.e. a text that does not belong to any corpus. Methods from this group rely on statistical \citep{luhn,rake,yake,matsuo}, or graph-theoretical features of a word in a text \citep{textrank,topicrank,positionrank,collabrank}; see section \ref{related}. The oldest method from this group is LUHN which selects sufficiently frequent content words of a text \citep{luhn}. One of the latest state-of-the-art methods from the third group is YAKE \citep{yake,yake!}.  

Here we focus on keyword extraction from literary works without supervision and corpus. One purpose of this task is to extract topical groups of keywords. We concentrate on well-known literary works because the confirmation of their keywords and topics should be available to practically anyone with a general education. (Sometimes manual keyword extraction and validation require specialist expertise.) Another reason to work with literary works is that regular keyword extraction schemes can be applied to discourse analysis \citep{keyness}. 

Our keyword extraction method belongs to the third group and is based on the specific spatial distribution of keywords in the text; see section \ref{method}. Systematic studies of the spatial distribution of words were initiated by \cite{zipf1945} and continued by \cite{yngve} and \cite{herdan}; see section \ref{related}. \cite{spatial5_ortuno,spatial1} suggested to employ the spatial distribution for detecting the keywords; see section \ref{related}. This suggestion was taken up by \cite{spatial2, spatial3, spatial4, spatial5}. 

However, several questions were open with these proposals. What are the best indicators for keywords based on spatial distribution-based methods? How do they compare to existing unsupervised single-text methods? Which keyword-extracting tasks can they help with?

We research these questions and to a large extent answer them. We use the spatial distribution of words for keyword detection, and our unsupervised and corpus-independent method is based on comparing the second (and sixth) moments of this distribution before and after a random permutation of words. By doing so, we capture two types of keywords: global and local. Global keywords are spread through the text and their spatial distribution becomes more homogeneous after a random permutation of words. By contrast, local keywords are found in particular parts of a text and clustered together. After a random permutation, their distribution becomes more homogeneous and this can be employed for keyword detection. Analyzing several classical texts, we saw that this structural difference between the keywords indeed closely relates to the content of the text; e.g., global and local keywords refer to (resp.) the main and secondary characters of the text. Thus, global keywords give the general idea of the text, whereas local keywords focus our attention on parts of the text. We note that the importance of global and local keywords was already understood in linguistics \citep{textual}, but no systematic method was proposed there for their detection. Related ideas on different types of keywords appeared in \cite{pedro}. 

Our method provides significantly better precision and recall of keyword extraction than several known methods including LUHN \citep{luhn} and YAKE \citep{yake,yake!}, KeyBERT \citep{bert}, KEA \citep{kea} and WINGNUS \citep{wingnus} (the first three methods are unsupervised, the latter two are supervised). We noted that despite its relative sophistication, for single-word keywords (i.e. not key phrases) extracted from literary works, YAKE provides results that always approximately coincide with those of LUHN, though it outperforms graph-based methods \citep{yake!} (we confirmed this point for texts studied here). Hence, we do not show comparison results with the latter group of methods. We also implemented for our texts another statistics-based method, RAKE \citep{rake}, to confirm that YAKE outperforms it. 

The advantage of our method is found via human annotators who determine if the extracted words are keywords based on their previous knowledge of classic literature texts in our database. There is moderate to substantial agreement between annotators. Additionally, we gave two indirect, but human-independent indications of the advantage of our method over the above methods. First, words extracted by our method have a longer length (in letters) than English content words on average. Therefore, we can infer indirectly that our method extracts text-specific words because it is known that the length of content words correlates with their average informativeness \citep{pnas}. Second, our method extracts more nouns. This is a proxy for keyword extraction since keywords are mostly nouns \citep{review4}. 

In contrast to existing methods, our method is able to find topics from the text, i.e. annotators were able to identify topical groups from a set of keywords extracted via our method. For the studied texts, serious topic extraction proved to be impossible with all alternative methods considered, including keyword extraction methods, as well as via several NLP topic modeling methods; see section \ref{topos}. Our method is also nearly language-independent, as verified using translations in three languages: English, Russian, and French. It is only for long texts that our keyword extraction method is more efficient. For short texts our method does apply, but its efficiency of keyword extraction is similar to those of LUHN and YAKE. Still, its power in extracting the textual topics remains even for short texts.   

To find out the limitations of our method, and to gain an understanding of what a keyword means conceptually, we aimed to relate keywords to the higher-order structures of texts, i.e. the fact that literary texts are generally divided into chapters. This was accomplished by developing a method of keyword extraction that is based on chapter division. Even though this method is less efficient than our main method, it is easier to use in practice (for texts that already have many chapters), and it has the potential for further development; see section \ref{chapters}. 

The rest of the paper is organized as follows. The next section reviews related work. In section \ref{method}, we discuss the main method analyzed in this work. Section \ref{anna} shows how the method applies to a classic and well-known text: {\it Anna Karenina} by L. Tolstoy. Section \ref{more} evaluates our results in various ways. The inter-annotator agreement is also discussed in this section. Section \ref{chapters} is devoted to the keyword extraction method that employs the fact that a long text is divided over sufficiently many chapters. The final section summarizes the discussion. Here we emphasize that we considered only single-word keywords, and the extension of our method to extraction of key phrases is an open problem. 

\section{Related work}
\label{related}

In discussing various keyword extraction methods, one must remember that they are not universally applicable: each task (e.g. information retrieval, information extraction, document classification, content analysis) requires its own methods. Keyword extraction methods are roughly divided into three groups: supervised, unsupervised but employing a text corpus, and unsupervised methods that apply to a single text. While in the context of the content analysis, we naturally focus on the last group, we shall also briefly review the two other groups. 

Supervised methods are discussed in \cite{gollapalli,genex,kpspotter,pollak}. For general reviews on such methods see \cite{review2,review3,review4,review5}. The supervision (training) stage normally demands a large training set with $>10^4$ documents. Hence such methods are prone to over-fitting and do not seem to be applicable for keyword extraction from a single literary work, though such applications are not excluded in principle and should be studied in the future. Some supervised approaches for keyword extraction employ linguistic-based handcrafted rules \citep{review4, textrank,hulth}, which however lack language independence. 

Unsupervised approaches include methods from statistics, information theory and graph-based ranking \citep{review3,review4,review5}. The most recent review of unsupervised approaches is \citep{akopian}. 
The best-known and widely used statistical approach is perhaps TF-IDF scoring function \citep{manning,Ando2005,tfidf1,tfidf2}. Ideas that are similar to TF-IDF were independently researched in corpus linguistics \citep{textual,keyness}. The method assumes that relevant words appear frequently in the given text and rarely in other texts in the corpus. Thus, the TF-IDF function relies on the existence of the corpus, i.e. it does not apply to a single text. 

Other unsupervised methods do apply to a single text. The first such method was proposed by \cite{luhn}. It takes frequent content words as keyword candidates, excludes both high-probable content words and low-probable content words, and selects the rest as keyword candidates 
\citep{luhn}. RAKE \citep{rake} and YAKE \citep{yake,yake!} are two other methods that employ statistical information and apply it to a single document (without a corpus). In particular, YAKE emerged as the current state-of-the-art keyword extraction algorithm.  

In graph-based methods \citep{textrank,positionrank, collabrank,rakun} a text is represented as a graph where nodes are words and relations between words are expressed by edges. Normally, better-connected nodes (e.g. as determined by PageRank algorithm) relate to keywords, though other network features such as betweenness and closeness were also studied in the context of keyword extraction \citep{pagerank,boudin}. These methods mainly differ by the principles used to generate edges between words \citep{topicrank}. Graph-based methods need only text information and hence are corpus-independent compared to TF-IDF. They can be semantically driven and agnostic of languages \citep{cake}.

KeyBERT is another unsupervised method of keyword extraction \citep{bert}. 
It inherits the pre-trained BERT model (Bidirectional Encoder Representations from Transformers) developed by Google that understands the context of words in a sentence by considering the words that come before and after it.
BERT is large language model that was trained on a large text corpus (including the entire English Wikipedia and the BookCorpus dataset) to learn language representations. 
Three recent keyword-extraction methods that employ language models are \cite{patternrank,entropyrank,unsupervised}.

Zipf and Fowler initiated systematic studies of the spatial distribution (or gap distribution) of words in texts \citep{zipf1945}. 
\cite{yngve} and \cite{herdan} noted that the gap distribution of words is far from random and that this fact can be employed in quantitative linguistics. A pertinent open question is how to characterize this randomness \citep{brainerd,zornig1,zornig2,pedro}. \cite{spatial5_ortuno} specifically applied the spatial distribution of words for detecting keywords in a single text, i.e. without training and without a corpus. In \citep{spatial5_ortuno}, the variance of the spatial distribution is used for finding clustered words that are related to keywords. Later works \citep{spatial1, spatial2, spatial3, spatial4} suggest several modifications that appear to improve the results. \cite{spatial1} proposed to combine Shannon's information measure with the spatial distribution, and studied the keyword distribution of {\it The Origin of Species} by Charles Darwin. Information-theoretic measures were also tried in \cite{spatial2, spatial3, spatial4}. An alternative metric for keyword extraction was proposed by \cite{spatial5}. However, this variety of methods employing spatial distribution was not applied to a sufficiently large database. Also, no systematic comparison was attempted with the existing methods of keyword extraction from a single text. It was also unclear to which specific keyword-extracting tasks these methods apply. These issues are researched below. 

\section{Method}
\label{method}

Below we discuss our method for keyword extraction (sections \ref{kkk}, \ref{spa2}), and describe implementation details; see sections \ref{lemma_ap} and \ref{lu}. Section \ref{spa1} introduces ideas on the example of spatial frequency, which shows interesting behavior, but does not result in productive keyword indicators.  

\subsection{Distribution of words: spatial frequency}
\label{spa1}

Our texts were lemmatized and freed from functional words (stop-words); see section \ref{lemma_ap} for details.
Let $w_{[1]},...,w_{[\ell]}$ denote all occurrences of a word
$w$ along the text. Let $\zeta_{i}$ denotes the number of words
(different from $w$) between $w_{[i]}$ and $w_{[i+1]}$; i.e. 
$\zeta_{\,i}+1\geq 1$ is the number of space symbols between 
$w_{[i]}$ and $w_{[i+1]}$. Define the first empirical moment for the distribution of
$\zeta_{\,i}+1$ \citep{yngve,halves}:
\BEA
\label{durnovo}
C_1[w]=\frac{1}{\ell-1}{\sum}_{i=1}^{\ell-1} \,(\zeta_{\,i}+1).
\EEA
Eq.~(\ref{durnovo}) is not defined for $\ell=1$, i.e. for words that occur only once; hence such words are to be excluded from consideration, i.e. they will not emerge as keywords. 

Note that $C_1[w]$ is the average period of the word $w$. Hence, the spatial frequency $\tau(w)$ can be defined via 
\citep{spatial5_ortuno, yngve, carpena, zano}:
\BEA
\label{hop}
\tau[w]\equiv 1/C_1[w].
\EEA
The smallest value $\frac{1}{N-1}$ of $\tau[w]$ is attained for $\ell=2$, where $w$ occurs as the first and last word of the text. The largest value $\tau[w]=1$ is reached when all instances of $w$ occur next to each other (strong clusterization of $w$). 

We compare $\tau[w]$ with the ordinary frequency $f[w]$ of word $w$:
\BEA
\label{ordinary}
f[w]=N_w/N, 
\EEA
where $N_w$ is the number of times $w$ appeared in the text ($N_w=\ell$), while $N$ is the full number of words in the text. Now $f[w]$ is obviously invariant under any permutation of words in the text. 

Note that $(\ell-1)(C_1[w]-1)$ equals to the number of words that differ from $w$ and occur between $w_{[1]}$ and $w_{[\ell]}$. Hence $\tau[w]$ will stay intact at least under any permutation of words in that part of the text which is located between 
$w_{[1]}$ and $w_{[\ell]}$. This class of permutation is sufficiently big for frequent words (in the sense of (\ref{ordinary})), where $w_{[1]}$ [$w_{[\ell]}$] occurs close to the beginning [end] of the text. Consequently, we expect that a random permutation of all words in the text will leave $\tau[w]$ nearly intact for frequent words: $\tau[w]\approx \tau_\perm[w]$. Indeed, we observed such a relation empirically. We also observed that there are many infrequent words for which $\tau[w]\gg \tau_\perm[w]$, i.e. such words are well-clustered (before permutation).

\begin{figure}[!ht]
\centering
    \includegraphics[width=0.8\columnwidth]{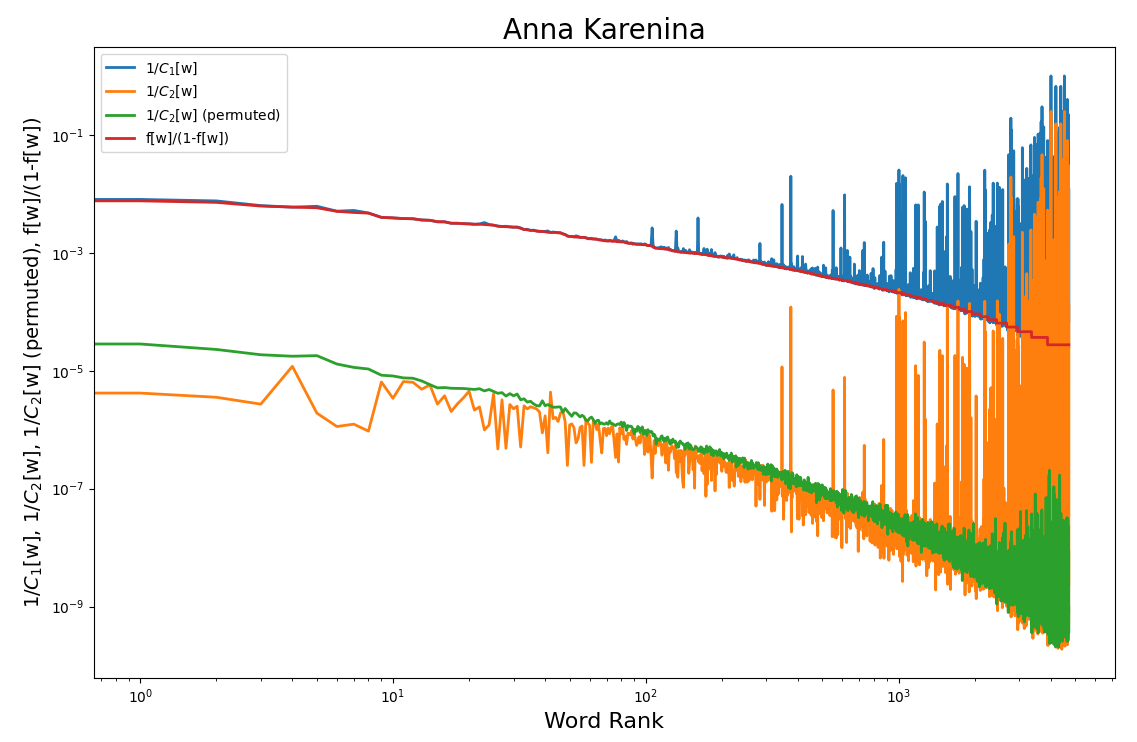}
    \vspace{2.5cm}
    \caption{For {\it Anna Karenina} by L. Tolstoy \citep{anna} we show space frequency $\tau[w]=1/C_1[w]$ and $1/C_2[w]$ {\it versus} word rank for all distinct words $w$ of the text; cf.~Eqs.~(\ref{durnovo}, \ref{3}). We also show two additional quantities: $1/C_2[w]=1/C_{2\,{\rm perm}}(w)$ after a random permutation of words in the text, and $f[w]/(1-f[w])$, where $f[w]$ is the frequency of $w$; see Eqs.~(\ref{dag}, \ref{ordinary}).
    Ranking of distinct words is done via $f[w]$, i.e. the most frequent word got rank 1, {\it etc}. It is seen that $C_{2\,\perm}[w]<C_{2}[w]$ holds for frequent words. Both $C_{2\,\perm}[w]<C_{2}[w]$ and $C_{2\,\perm}[w]>C_{2}[w]$ hold for less frequent words. Not shown in the figure: a random permutation of the words in the text leaves $\tau[w]$ unaltered for frequent words, while $\tau[w]$ generically increases for less frequent words (clusterization); cf.~Eq.~(\ref{dag}). 
        }
\label{annafig}
\end{figure}

\begin{figure}[!ht]
\centering
    \includegraphics[width=0.8\columnwidth]{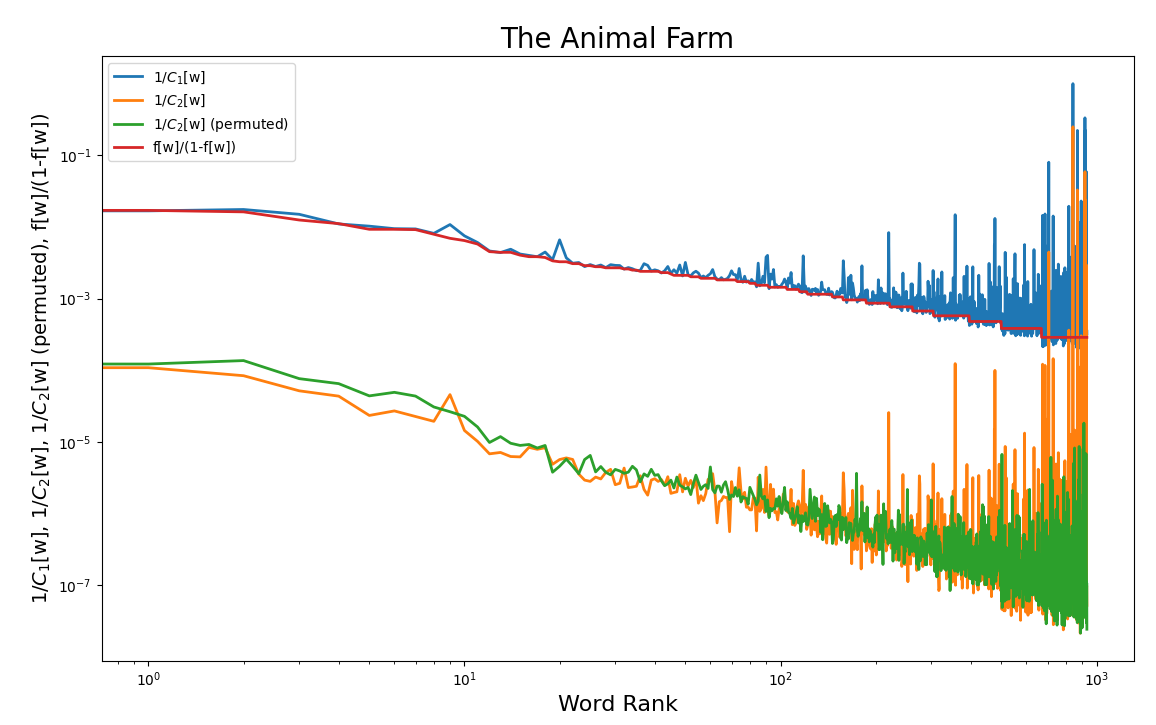}
    \vspace{2.5cm}
    \caption{For {\it Animal Farm} (AF) by G. Orwell we show the same quantities as for  
    {\it Anna Karenina} (AK) in Fig.~\ref{annafig} (also the same notations). AK is 11.6 times longer than AF; see Table~\ref{tab_gogo}. Some differences between these texts are as follows. Inequality $C_{2\,{\rm perm}}(w)<C_2(w)$ holds for a lesser number of frequent words in AF compared with AK. Domain $C_{2\,{\rm perm}}(w)<C_2(w)$ and $C_{2\,{\rm perm}}(w)>C_2(w)$ are well-separated in AK, and not so well-separated in AF. For AF, relation (\ref{dag}) can be violated for some infrequent words.  }
\label{farmfig}
\end{figure}

These relations can be made quantitative by noting for frequent words the following implication of the above invariance. 
Aiming to calculate $\tau_\perm[w]$ for a given frequent word $w$ we can employ the Bernoulli process of random text generation, assuming that $w$ is generated independently from others, with probability of $w$ (not $w$) equal to $f[w]$ ($1-f[w]$); see (\ref{ordinary}). For spatial interval $s$ between the occurrences of $w$ the Bernoulli process produces the geometric distribution ${p}(s)=(1-f[w])^sf[w]$, where for sufficiently long texts we can assume that $s$ changes from $0$ to $\infty$, and $\sum_{s=0}^\infty p(s)=1$. We emphasize that this model is not precise for a random permutation in texts, but it turns out to be sufficient for estimating $\tau_\perm[w]$. 
The mean of this distribution is 
\BEA
f[w]{\sum}_{s=0}^{\infty}s(1-f[w])^s={(1-f[w])}/{f[w]}.
\label{geom}
\EEA
The inverse of (\ref{geom}) estimates $\tau_\perm[w]$ for frequent words $\tau_\perm[w]\simeq f[w]/(1-f[w])$. On the other hand, we have $\tau[w]\simeq f[w]/(1-f[w])$ for frequent words; see Figs.~\ref{annafig} and \ref{farmfig}. Two of the most famous world literature texts are described in these figures. Fig.~\ref{annafig} refers to {\it Anna Karenina} by L. Tolstoy (the total number of words $N\approx 3.5\times 10^5$), and Fig.~\ref{farmfig} refers to {\it Animal Farm} by G. Orwell ($N\approx 3\times 10^4$). The length difference between the two texts is reflected in the difference between $\tau[w]$ and $f[w]/(1-f[w])$. Fig.~\ref{annafig} shows that relation 
\BEA
f[w]/(1-f[w])\lesssim\tau[w],
\label{dag}
\EEA
holds for the majority of words with approximate equality for frequent words. In Fig.~\ref{farmfig}, relation (\ref{dag}) holds for frequent words, but is violated for some not-frequent words. For both figures, we see that $\tau[w]$ can be significantly larger than $f[w]/(1-f[w])$ for certain non-frequent words, indicating that the distribution of such words is clustered. We checked that there are not many keywords among such words, i.e. the magnitude of $\tau[w](1-f[w])/f[w]$ is not a productive indicator for the keywords. More refined quantities are needed to this end. 

\subsection{ The keyword extraction method: the second moment of the spatial distribution } 
\label{kkk}

Given Eq.~(\ref{durnovo}), let us define the second moment of the spatial distribution for word $w$ 
\begin{equation}
\label{3}
C_2[w]=\frac{1}{\ell-1}{\sum}_{i=1}^{\ell-1} \,(\zeta_{\,i}+1)^2.
\end{equation}
$C_2[w]$ is not invariant to those word permutations that left invariant $C_1[w]$; cf.~the discussion before (\ref{dag}). 
Figs.~\ref{annafig} and \ref{farmfig} show that for sufficiently frequent words $w$, $C_2[w]$ decreases after a random permutation.
Indeed, frequent words are distributed in the text inhomogeneously. A random permutation makes this distribution more homogeneous and hence makes $C_{2\, \perm}[w]<C_2[w]$ for frequent words. For this conclusion, we need the second (or higher-order) moment in (\ref{3}). Appendix \ref{cuba} presents a numerical illustration of this effect, and also illustrates that $C_{1}$ does not catch it. 

The situation changes for less frequent words: now it is possible that for some words non-frequent words we get $C_{2\, \perm}[w]>C_2[w]$; see Appendix \ref{cuba} for examples. Those words are clustered in the original text, while after a random permutation, their distribution is more homogeneous; see Figs.~\ref{annafig} and \ref{farmfig}. For a long text {\it Anna Karenina}, the words where $C_{2\, \perm}[w]$ is noticeably larger than  $C_2[w]$ appear at rank $\approx 300$ (the rank is decided by frequency (\ref{ordinary})); see Fig.~\ref{annafig}. There is no such a sharp threshold value for a shorter text {\it Animal Farm}, as Fig.~\ref{farmfig} shows. Using Eq.~(\ref{3}), we define
\begin{equation}
\label{37}
{A[w]}=\frac{C_{2\,\perm}[w]}{C_2[w]},
\end{equation}
where $C_{2\,\perm}[w]$ is calculated via Eq.~(\ref{3}) but after a random permutation of all words of the text.

When checking the values of $A[w]$ for all distinct words of several texts, our annotators concluded that sufficiently small and sufficiently large values of $A[w]$ in Eq.~(\ref{37}), 
\BEA
\label{40}
&& A[w]\leq \frac{1}{5},\\
&& A[w]\geq 5,
\label{41}
\EEA
can be employed for deducing certain keywords of the text. Eq.~(\ref{40}) extracts global keywords of the text, i.e. keywords that go through the whole text. Eq.~(\ref{41}) refers to local keywords, i.e. those that appear in specific places of the text. In Fig.~\ref{annafig} they are seen as local maxima of $1/C_2[w]$. Local keywords are naturally located in the domain of infrequent words.

Taking in Eq.~(\ref{40}) a smaller threshold values 
\BEA
\frac{1}{5}\leq A[w]\leq \frac{1}{3}, 
\label{ade}
\EEA
leads to selecting a group of lower-frequency global keywords. Below we shall refer to Eq.~(\ref{40}) and Eq.~(\ref{ade}) as (resp.) strong and weak cases. 

Relations of Eq.~(\ref{40}) and Eq.~(\ref{41}) with (resp.) global and local keywords make intuitive sense. As we checked in detail, spaces between global keywords assume a broad range of values. This distribution becomes more uniform after the random permutation, hence the the second moment decreases; cf.~Eq.~(\ref{40}). Local keywords refer to infrequent words, are localized in a limited range of text, and are clustered. Hence a random permutation increases the second moment; cf.~ Eq.~(\ref{41}).

Let us comment on the choice of parameters in Eqs.~(\ref{40}, \ref{ade}). This choice was taken as empirically adequate for {\it Anna Karenina}, i.e. it led to extracting sufficiently many local and global keywords. (Other choices led to fewer global and/or local keywords.) After that, it was applied for all long texts [see Table~\ref{tab_gogo}] and led to adequate results.

\comment{The simplest method of looking for outliers in a random sample $\mathcal{X}=(x_1,...,x_N)$ of numerical data; see \cite{taras} for a review on similar and more advanced methods. Let the distinct elements of $\mathcal{X}$ be $\{\chi_k\}_{k=1}^n$, so that the frequency of $\chi_k$ in $\mathcal{X}$ is $p_k$. We can now decide whether a given element $x_\ell$ in $\mathcal{X}$ is an outlier. More precisely, $x_\ell$ is an outlier at the validity level $n$ ($n=3,4,5$), if $\sigma/|x_\ell-\bar{x}|<1/n$, where $\bar{x}=\sum_{k=1}^np_k\chi_k$ and $\sigma^2=\sum_{k=1}^np_k(\chi_k-\bar x)^2$ are (resp.) the mean and variance. Comparing $\sigma/|x_\ell-\bar{x}|$ with $A(w)$ in Eq.~(\ref{37}) we see that these quantities are similar to each, i.e. $A(w)$ is also a ratio of the randomized variance to the actual distance from the mean. Hence conditions (\ref{40}, \ref{ade}) resemble methods applied for outlier detection confirming an intuitive point that global keywords relate to outliers. }

As our method relies on random permutations, our results are formally dependent on the realization of these permutations. (Random permutations of words were generated via Python's numpy library; see  Appendix \ref{external}.) Such a dependence is weak: we noted that only a few keywords change from one realization to another. However, we cannot avoid random permutations; see section \ref{final} for further discussion. 

\comment{Let us illustrate this point with the geometrical distribution proposed in \cite{yngve,herdan,spatial1,spatial3}. In a long text, the probability $p(s)$ for spaces (gaps) $\zeta_{\,i}(w)=s$ [c.f.~Eq.~(\ref{durnovo})] of a word $w$ after a random permutation is asymptotically geometrical, $p(s)=(1-f(w))f^s(w)$ ($s\geq 0$) provided that its frequency $f(w)$ is big enough; cf.~Eq.~(\ref{ordinary}). But for the majority of keywords this asymptotic limit is not reached, since $f(w)$ is not big. }

\comment{
\cite{spatial1} suggested to calibrate the variance by taking the ratio of standard deviations of original and random texts justifying this approach by the fact that the variance contains the frequencies of words, hence the need to eliminate this dependence. Here the problem is to find a good model for a random version of original text. The authors suggest to use geometric distribution for $\zeta_{\,i}-$s where the probability of success (occurrence of a specific word $f(w_r)$) is taken as its frequency in original text $f(w_r)$. We see that this intuition does not really work. Alternative versions of a randomly permuted text are described also by \cite{spatial3}. }

\subsection{Modification of the method for shorter texts}
\label{spa2}

Criteria (\ref{40}, \ref{41}) based on $A(w)$ from Eq.~(\ref{37}) are not sufficiently powerful for discriminating between the keywords and ordinary words in sufficiently short texts; e.g., in {\it Animal Farm} depicted on Fig.~\ref{farmfig}. 
We found two modifications of the method that apply to short texts. The first option is to look at local maxima and minima of $A(w)$. The second, better option is to modify the order of the moment in Eq.~(\ref{3}). Instead of the second moment in Eq.~(\ref{3}) we employed the sixth moment
\BEA
\label{66}
C_6[w]=\frac{1}{\ell-1}{\sum}_{i=1}^{\ell-1} \,(\zeta_{\,i}+1)^6.
\EEA
This modification leads to an indicator (\ref{sharik}), which is more susceptible to inhomogeneity and clustering. 
Now $A_6[w]$ is defined analogously to Eq.~(\ref{37}), but via Eq.~(\ref{66}), 
\BEA
\label{dog}
A_6[w]=\frac{C_{6\,\perm}[w]}{C_6[w]},
\EEA
and for extracting keywords we can apply [cf.~Eqs.~(\ref{40}, \ref{41})]:
\BEA
A_6[w]\leq \frac{1}{3}, ~~~ A_6[w]\geq 3.
\label{sharik}
\EEA
The utility of Eqs.~(\ref{dog}, \ref{sharik}) was determined for the short text {\it Animal Farm}, and then applied for all other short texts; see Table~\ref{tab_gogo}.

\subsection{Lemmatization of texts}
\label{lemma_ap}

English texts were preprocessed using WordNetLemmatizer imported from nltk.stem; see Appendix \ref{external}. This library looks for lemmas of words from the WordNet Database. The lemmatization uses corpus for excluding stop-words (functional words) and WordNet corpus to produce lemmas. WordNetLemmatizer identifies the intended part of speech and meaning of a word in a sentence, as well as within the larger context surrounding that sentence, such as neighboring sentences or even an entire text. We applied this lemmatization algorithm on nouns, adjectives, verbs and adverbs to get maximal clean up of the text. Any stemming procedure will be inappropriate for our purposes of extracting keywords, since stemming may mix different parts of speech. 

For inflected languages (e.g. Russian), the lemmatization rules are more complex. 
For French and Russian texts we used (resp.) lemmatizers LEFFF and pymystem3; see Appendix \ref{external}. 

\subsection{Implementation of LUHN and YAKE}
\label{lu}

Here we briefly discuss how we implemented Luhn's method (LUHN) for keyword extraction \citep{luhn}. The method starts with ranking the distinct words of the text according to their frequency (more frequent words got a larger rank): $\{f_r\}_{r=1}^n$, where $f_r$ is the word frequency and $r$ is its rank. Next, one cuts off the high-frequency and low-frequency words and selects the remaining words $\{f_r\}_{r=r_{\rm min}}^{r_{\rm max}}$ as candidates for keywords. Hence the method amounts to selecting the above cut-offs $r_{\rm min}$ and $r_{\rm max}$: the high-frequency words are to be omitted because there are many stop-words there that normally are not considered as keywords. Low-frequency words are to be omitted since they are not relevant to the semantics of the text. Once we already skipped functional words from our texts, we did not apply the high-frequency cut-off, i.e. we take $r_{\rm min}=1$. For the low-frequency threshold, we employed a hypothesis that the essence of Luhn's method is related to Zipf's law, i.e. to the law that fits the rank-frequency curve of distinct words of a text to a power law \citep{zipf1945,zipf_pre}. It is known that the power-law fitting works approximately till the rank $r_{10}$ so that for $r\geq r_{10}$ the number of words having the same frequency $f_r$ is $10$ or larger \citep{zipf_pre}. For $r>r_{10}$ the rank-frequency curve starts to show steps, that cannot be fitted to a single power-law curve, i.e. the proper Zipf's law becomes ill-defined for $r>r_{10}$ \citep{zipf_pre}. Hence we selected the rank $r_{\rm max}=r_{10}$. This choice shows reasonable results in practice. 

Both LUHN and our method are based on a unique idea with a straightforward implementation. In contrast, YAKE incorporates various tools and ideas along with numerous empirical formulas \citep{yake,yake!} (The same, but with a lesser degree holds for RAKE \citep{rake}). In particular, YAKE includes: textual context of candidate keywords; sentence structure; sliding windows for keyword selection; statistics of n-grams; non-trivial (and multi-parametric) scoring process; {\it etc}. We worked with the version of YAKE that was implemented via a Python package; see Appendix \ref{external}. We employed YAKE both with and without pre-processing of texts. In the second case, YAKE was better at extracting capitalized proper nouns (such words are frequently keywords). Otherwise, its performance did not change much. 
This advantage of YAKE is due to a specific tool implemented in it: YAKE looks for capitalized words which do not appear in the beginning of a sentence. Such a tool is easy to implement in any keyword searching method, but we avoided doing that, since we are interested in checking ideas behind Eqs.~(\ref{3}--\ref{sharik}). Therefore, we mostly discuss YAKE's outcomes after pre-processing. 

\comment{Lida, your evaluation for not-pre-processed Anna Karenina in YAKE led to the precision 23.4\% (there is a saved file on that). This was mostly due to proper nouns that start with a capital letter, one of many tricks implemented in YAKE. }

\section{Keywords extracted from {\it Anna Karenina}}
\label{anna}

The above keyword extraction method was applied to several texts of classic literature; see Table~\ref{tab_gogo}. (Our data for texts from Table~\ref{tab_gogo} is freely available at \url{https://github.com/LidaAleksanyan/keywords_extraction_data/tree/master}, while our codes are available at \url{https://github.com/LidaAleksanyan/spatial_keyword_extraction}.) Among them, we choose one of the most known works of classic literature, {\it Anna Karenina} by L. Tolstoy, and analyze in detail the implications of our method in extracting and interpreting its keywords. The evaluation of extracted keywords was done by annotators with expert knowledge of classic Russian literature and specifically works by Tolstoy. The agreement between annotators is moderate to substantial; see Table~\ref{tab_inter}. 

\subsection{Comparison with known methods of keyword extraction and language independence}

Using {\it Anna Karenina} \citep{anna}, we compared our approach discussed in section \ref{kkk} with two well-known methods that also apply to a single text (i.e. do not require corpus): LUHN and YAKE; see section \ref{lu}. 

282 words were extracted via each method, and the keywords were identified. Tables~\ref{tab_gogo} and \ref{tab_lang} show that for three languages (English, Russian, French) our method is better in terms of both precision and recall; see Appendix \ref{prerecall} for a reminder of these concepts. The relatively poor performance of YAKE and LUHN can be explained by their focus on relatively short content words that are not likely to be keywords. We quantified this by calculating the mean number of letters in each set of 282 words. For our method, LUHN and YAKE the mean is (resp.) 6.95, 5.43, and 5.5; cf.~the fact that the average number of letters in English content word is 6.47 (for stop word it is 3.13) \citep{miller}.

The three methods have scores for words. In LUHN and our method, the score coincides with the word frequency. 
For YAKE the scores are described by \cite{yake,yake!}.
However, for LUHN and YAKE the score did not correlate with the feature of being keyword. For our method it certainly did, i.e. by selecting only high-score words we can significantly enlarge the percentage of keywords compared to what is seen in Table~\ref{tab_gogo}. These two facts (low density of keywords plus no correlation with their score) make it impossible to extract topical groups of keywords via LUHN and YAKE; cf.~the discussion after Eq.~(\ref{41}). 

Another comparison criterion between the three methods is the amount of nouns in words that were not identified as keywords. 
Indeed, once keywords are predominantly nouns, a method that extracts more nouns (e.g., more nouns in candidate words that were not identified as keywords) has an advantage. In this respect, our method fares better than both LUHN and YAKE; see Table~\ref{tab_lang}. 

Table~\ref{tab_lang} also addresses the language independence of the three methods that were studied in three versions (English, Russian, and French) of {\it Anna Karenina}. Our method performs comparably for English and Russian, which are morphologically quite distinct languages. For French the performance is worse, but overall still comparable with English and Russian. Altogether, our method applies to different languages. This confirms an intuitive expectation that spatial structure features embedded into Eqs.~(\ref{3}--\ref{sharik}) are largely language-independent. 

\subsection{Topical groups extracted via extracted keywords}

{\it Anna Karenina} features more than a dozen major characters and many lesser characters. Annotators separated keywords into 9 topical groups: \1 proper names of major characters; \2 proper names of secondary characters; \3 animal names; \4 trains and railway; \5 hunting; \6 rural life and agriculture; \7 local governance (zemstvo); \8 nobility life and habits; \9 religion; see Table~\ref{tab_groups}. 

The names of these characters are keywords, because they inform us about the character's gender ('anna' {\it vs.} 'vronsky'), age ('alexandrovitch' {\it vs.} 'seryozha') and the social strata; e.g. 'tit' {\it vs.} 'levin'. Proper nouns provide additional information due to name symbolism employed by Tolstoy; e.g. 'anna'='grace'; 'alexey'='reflector'; 'levin'='leo' is the {\it alter ego} of Tolstoy \citep{gustafson}. 

All the main character names came out from our method as strong global keywords holding condition $A[w]\leq \frac{1}{5}$ in Eq.~(\ref{40}): 'levin', 'anna', 'vronsky', 'kitty', 'alexey', 'stepan', 'dolly', 'sergey'; see Table~\ref{tab_groups} for details. Many pertinent lesser characters came out as local keywords, as determined via Eq.~(\ref{41}); e.g. 'vassenka', 'golenishtchev', 'varvara'; see Table~\ref{tab_groups}. Important characters that are not the main actors came out as weak global keywords, e.g. 'seryozha', 'yashvin', 'sviazhsky'.

The novel is also known for its animal characters that play an important role in Tolstoy's symbolism \citep{gustafson}. Our method extracted as local keywords the four main animal characters: 'froufrou', 'gladiator' 'laska', 'krak'. Trains are a motif throughout the novel (they symbolize the modernization of Russia), with several major plot points taking place either on passenger trains or at stations in Russia \citep{anna,gustafson}. Our method extracted among the global keywords 'carriage', 'platform' and 'rail'. Hunting scenes are important in the novel depicting the life of Russian nobility. Accordingly, our method extracted keywords related to that activity: 'snipe', 'gun', 'shoot'. Two major social topics considered in the novel are local democratic governance (Zemstvo) and the agricultural life of by then mostly rural Russia. For the first we extracted keywords:  'district', 'bailiff', 'election' {\it etc}. And for the second: 'mow', 'lord', 'acre', {\it etc}. A large set of keywords are provided by Russian nobility's living and manners, including their titles, professions and habits; see Table~\ref{tab_groups}. Religion and Christian faith is an important subject of the novel. In this context, we noted keyword 'Lord', 'priest', 'deacon'; see Table~\ref{tab_groups}. Finally, a few words stayed out of these topical groups but was identified as keywords: 'lesson', 'crime', 'cheat', 'salary', 'irrational', 'law', 'skate', 'tribe'. 

\comment{ 'princess', 'prince', 'countess', 'madame', 'officer', 'telegram', 'bedroom', 'nobleman', 'article', 'professor', 'sportsman', 'armchair', 'director', 'minister', 'captain', 'matrona', 'duel', 'mazurka', 'villa', 'bridgeroom', 'photograph', 'architect', 'pa', 'mais', 'bedchamber', 'opium', 'palazzo', 'crown', 'adultery', 'musical'.}

\begin{table}\caption{Analyzed long texts: {\it Anna Karenina}, {\it War and Peace, part I}, and {\it War and Peace, part II} by L. Tolstoy; {\it Master and Margarita} by M. Bulgakov; {\it Twelve Chairs} by I. Ilf and E. Petrov; {\it The Glass Bead Game} by H. Hesse; {\it Crime and Punishment} by F. Dostoevsky. Shorter texts: {\it The Heart of Dog} by M. Bulgakov; {\it Animal Farm} by G. Orwell. {\it Alchemist} by  P. Coelho. Next to each text, we indicate the number of words in it, stop-words included. \\ For long texts we extracted for each text the same number of $\approx 300$ potential keywords via each method: our method (implemented via Eqs.~(\ref{40}, \ref{41}, \ref{ade})), LUHN and YAKE. The numbers below are percentages, i.e. $15.6= 15.6\%$.
For each text, the first percentage shows the values of precision (Prec.), i.e. the fraction of keywords which were identified as keywords by human annotators. The second percentage shows recall (Rec.): the fraction of keywords that the methods were able to extract compared to ground truth keywords; see (\ref{punjab}). The third percentage shows the F1 score; see (\ref{gloria}).\\
For short texts, we extracted via each method $\sim 100$ words. Our method was implemented via Eq.~(\ref{sharik}); only the precision is shown. For longer texts, our method provides sizable advantages compared with LUHN and YAKE. For shorter texts the three methods are comparable (the values for recall are not shown). }
\begin{tabularx}{1\textwidth} 
{ 
  | >{\raggedright\arraybackslash} X|
  | >{\raggedright\arraybackslash} X|
  | >{\raggedright\arraybackslash} X|
  | >{\raggedright\arraybackslash} X|}
 \hline
 Method & LUHN  & YAKE & Our method\\
 \hline
 Text                             & {\rm Prec.}  ~~~~~~{\rm Rec.}~~~~~~{\rm F1} & {\rm Prec.}~~~~~~{\rm Rec.}~~~~~{\rm F1} & {\rm Prec.}~~~~~~{\rm Rec.}~~~~~{\rm F1} \\
 \hline\hline
 {\it Anna Karenina} (349762) & 15.6  ~~~~~~26.5~~~~~~19.6 & 15.6~~~~~~25.9~~~~~19.5 & 55.6~~~~~~91.2~~~~~69.1 \\
  \hline
 {\it War and Peace, part I} (142254) & 22.7  ~~~~~39.1~~~~~~29.0 & 22.7~~~~~39.1~~~~~~29.0 & 42.9~~~~~81.7~~~~~~56.3 \\
\hline
 {\it War and Peace, part II} (128146) & 28.1~~~~~~45.1~~~~~~34.5 & 26.0~~~~~45.1~~~~~~33.8 & 51.0~~~~~69.7~~~~~~58.9 \\
 \hline
 {\it Master and Margarita} (145286) & 18.5~~~~~~27.5~~~~~22.5 & 18.2~~~~~30.4~~~~~22.6 & 55.9~~~~~92.6~~~~~69.8 \\
 \hline
 {\it Twelve Chairs} (102485) & 20.5~~~~~~36.7~~~~~~25.9 & 20.5~~~~~38.3~~~~~~26.3 & 38.3~~~~~67.5~~~~~~50.2 \\
 \hline
 {\it The Glass Bead Game} (192311) & 17.7 ~~~~~~36.5~~~~~~24.1 & 19.3~~~~~39.7~~~~~~25.7 & 33.9~~~~~67.9~~~~~~45.3 \\
 \hline
 {\it Crime and Punishment} (203505) & 16.3~~~~~~39.3~~~~~~34.0 & 18.5~~~~~42.6~~~~~~26.3 & 29.6~~~~~81.9~~~~~~43.9 \\
 \hline\hline
 {\it The Heart of Dog} (34950)   & 26.2  & 24.3 & 26.2 \\
 \hline
 {\it Animal Farm} (30037)   & 49.6  & 47.3 & 45.8 \\
 \hline
 {\it Alchemist} (39004) & 32.5  & 33.0 & 29.5 \\
 \hline
\end{tabularx}
\label{tab_gogo}
\end{table}

\begin{table*}\caption{The values of Cohen's kappa (\ref{kappa1}) for the agreement in keyword extraction tasks between two annotators for three different texts; see section \ref{inter2}. The keyword extraction employed the method discussed in Table~\ref{tab_gogo} and section \ref{kkk}. Results for global and local keywords are shown separately. It is seen that the agreement is better for global keywords. A possible explanation is that the annotators do not focus on text details. }
\begin{tabular}{ |c|| c| c| }
 \hline
 Text/type of keyword & global & local \\ 
 \hline
{\it Animal Farm} & 0.68 & 0.40  \\  
 \hline
{\it Alchemist} & 0.83 & 0.53   \\
 \hline
{\it Master and Margarita} & 0.78  & 0.65 \\
 \hline
\end{tabular}
\label{tab_inter}
\end{table*}

\begin{table*} 
\caption{Comparison of 3 different keyword extraction methods for English, Russian, and French versions of {\it Anna Karenina}. Percentages for keywords indicate the precision [cf.~Table~\ref{tab_gogo}], while ``nouns'' means the percentage of nouns in candidate words that were not identified as keywords. For all cases, our method fares better than LUHN and YAKE.  }
\begin{tabularx}{1\textwidth} 
{ 
  | >{\raggedright\arraybackslash}X |
  | >{\raggedleft\arraybackslash}X 
  | >{\raggedleft\arraybackslash}X |
  | >{\raggedleft\arraybackslash}X 
  | >{\raggedleft\arraybackslash}X |
  | >{\raggedleft\arraybackslash}X 
  | >{\raggedleft\arraybackslash}X |
  }
 \hline
 Method & English keywords & English nouns & Russian keywords & Russian nouns & French keywords & French nouns    \\
 \hline
 LUHN & 15.6 \% & 54 \%  & 14.1\%  & 51.1\%  & 19.2\%  & 62.3 \% \\
\hline
 YAKE & 15.6 \% & 55 \%  & 14.8\%  & 49.2\%  & 18\% & 60 \%  \\
\hline
Our~method & 55.6 \%  & 82 \% & 55\%   & 86.2\%   & 50.7\%  & 77.3\%  \\
\hline
\end{tabularx}
\label{tab_lang}
\end{table*}

\begin{table*}\caption{Words of {\it Anna Karenina} extracted via our method. For global keywords strong and weak cases mean (resp.) that the words $w$ were chosen according to $A(w)\leq \frac{1}{5}$ and $\frac{1}{5}\leq A(w)\leq \frac{1}{3}$; cf.~Eqs.~(\ref{40}, \ref{ade}). Local keywords were chosen according to $A(w)\geq {5}$; see Eq.~(\ref{41}). For each column, the words were arranged according to their frequency Eq.~(\ref{ordinary}). Keyword classes are denoted by upper indices; see details in the text. 
The last group \0 denotes words that were identified as keywords but did not belong to any of the above groups. Words without the upper index were not identified as keywords. }
\begin{tabularx}{\textwidth} 
{ 
  | >{\raggedright\arraybackslash}X 
  | >{\raggedright\arraybackslash}X |
   | >{\raggedright\arraybackslash}X | }
 \hline
 Global keywords strong cases & 
  Global keywords weak cases & 
   Local keywords \\ \hline
  levin\1, anna\1, vronsky\1, kitty\1, alexey\1, stepan\1, alexandrovitch\1, arkadyevitch\1, dolly\1, sergey\1, ivanovitch\1, peasant\6, darya\1, alexandrovna\1, varenka\1, lidia\1, death, ivanovna\1, laborer\6, mow\6, district\7, stahl\1, bailiff\5, gun\5, snipe\5, plough\6, rain, lesson\0, lord\9, acre\6, platform\4, natalia\1, built, rich, overlook, river, crime\0, rail\6, relate, throb, contrast, puzzle, cheat\0, oppress, irrational\0 
   &
love, princess\8, brother, carriage\4, horse\8, prince\8, doctor\8, countess\8, madame\8, sviazhsky\1, land\6, seryozha\1, konstantin\1, picture, oblonsky\1, nikolay\1, agafea\2, katavasov\2, grass\6, yashvin\1, shoot\5, mihalovna\2, officer\8, box, marshal\7, mare\6, priest\9, tree\6, forest\6, laska\3, law\0, landowner\6, realize, scythe\6, telegram\8, meadow\6, bedroom\8, argument, sledge, nobleman\8, paint, article\8, professor\8, scream, sky, trap, birch\6, cow\6, debt\0, rent, punish, sow\6, annushka\2, lightly, sportsman\8, myakaya\2, invalid, smart, parent, vividly, maman\8, institution\7, stable, distance, salary\0, educate, firm, skirt, mahotin\2, reconciliation, yellow, plump, childrens, tatar\2, outer, steward\8, cousin, loathsome, sharp, splash, armchair\8, understands, coarse, quicken, grace, delicious, director\8, unseen, selfpossession, cheese, rate, physically, timidity, tucked, reassure, sunday, compartment, frost, minister\8, won, king, repent, clock, wage, shock, uncertain, deliver, cream, silently, monday, captain\8, shaft\6, matrona\8, strictly, original 
&
 vassenka\2, golenishtchev\2, election\7, skate\0, varvara\2, pyotr\2, lizaveta\2, landau\2, petrovna\2, gladiator\3, metrov\2, tit\2, vote\7, froufrou\3, ryabinin\2, volunteer\8, nevyedovsky\2, duel\8, scandal\8, tribe\0, snetkov\2, lukitch\2, mower\6, deacon\9, native, korsunsky\2, hospital, remote, mazurka\8, pilate\0, sappho\0, villa\8, rival, reed\6, bridegroom\8, krak\3, merkalova\2, vorkuev\2, photograph\8, yegor\2, mitya\2, kapitonitch\2, architect\8, intensely, elect\7, golenishtchevs\2, pa\8, birthday, trousseau\8, transition, chalk, potato\6, kritsky\2, ergushovo\6, katya\2, weep, sympathetic, repair, mais\8, seryozhas\2, ballroom\8, classical, vozdvizhenskoe\6, technique, bedchamber\8, opium\8, penetrate, tchirikov\2, rider, palazzo\8, crown\8, remove, miracle, intolerable, turk\2, ballot\7, custom, nevsky\8, adultery\8, ditch, musical \\
\hline
\end{tabularx}
\label{tab_groups}
\end{table*}

\begin{table*}\caption{Here we discuss topical groups extracted from a short text. {\it Heart of Dog} by M. Bulgakov is a known satirical novella that shows the post-revolutionary Moscow (first half of the 1920s) under social changes, the emergence of new elites of Stalin's era, and science-driven eugenic ideas of the intelligentsia. Eventually, the novella is about the life of a homeless dog Sharik (a standard name for an unpedigreed dog in Russia) picked up for medical and social experiments. The majority of keywords below were not even extracted via LUHN and/or YAKE.}
\begin{tabularx}{1\textwidth} 
{ 
  | >{\raggedright\arraybackslash} X   |
  | >{\raggedright\arraybackslash} X|}
 \hline
 Topical group & Keywords\\
  \hline
 Canine features& 'dog', 'sharikov', 'salami', 'sharikovs', 'bite', 'cracow', 'scald', 'animal', 'sharik', 'cat', 'bitten', 'canine', 'phewphew', 'cur', 'claw', 'bitch', 'mange', 'shaggy', 'phew', 'paw', 'bark', 'wild', 'biting', 'oooo' \\
 \hline
 Medical terms & 'skull', 'camphor', 'weight', 'temperature', 'method', 'stitch', 'pulse', 'organism', 'injection', 'laboratory', 'sore', 'needle', 'scissors', hospital', 'respiratory', 'gauze', 'adrenaline', 'clinic', 'doze', ‘heal', 'transplant', 'phonograph', 'hypothesis', 'organism', 'nostril', 'injection',  'subdepartment', 'department', 'laboratory', 'hospital', 'rejuvenation', 'throat', 'scholar', 'brow', 'cheek', 'lip', 'strip', 'experiment', 'forehead', 'hormone', 'breast', 'science', 'hypophysis', 'brain' \\
 \hline
Revolution & 'proletariat', 'terror', 'kautsky', 'council', 'bourgeois', 'proletarian', 'revolution', 'war', 'worker', 'engels', 'pest', 'revolver', 'social', 'chairman', 'committee' \\
 \hline
 Moscow & 'moscow', 'kalabukhov', 'blizzard', 'bolshoi', 'mosselprom', 'nikitins', 'prechistenka', 'swearword' \\
 \hline
\end{tabularx}
\label{hd1}
\end{table*}

\comment{

\begin{table*}\caption{Comparison of 3 different keyword extraction methods for English version of {\it Anna Karenina}}
\begin{tabularx}{1\textwidth} 
{ 
  | >{\raggedright\arraybackslash}X 
  | >{\raggedleft\arraybackslash}X |
  | >{\raggedleft\arraybackslash}X |
  | >{\raggedleft\arraybackslash}X |
  | >{\raggedleft\arraybackslash}X |
  | >{\raggedleft\arraybackslash}X |}
 \hline
 Method & 
Number of extracted words & keywords/words & nouns(in non keyword set)/words & verbs(in non keyword set)/words & adj+adverbs(in non keyword set)/words\\
 \hline
 LUHN & 
282 & 10.6 & 42.2 & 34.4 & 12.8 \\
\hline
 YAKE &
 282 & 11.3 & 41.8 & 32.3 & 14.6 \\
\hline
OUR METHOD &
282 & 31.6 & 48 & 9.4 & 11 \\
\hline
\end{tabularx}
\label{table_english}
\end{table*}

\begin{table*}\caption{Comparison of 3 different keyword extraction methods for Russian version of {\it Anna Karenina}}
\begin{tabularx}{1\textwidth} 
{ 
  | >{\raggedright\arraybackslash}X 
  | >{\raggedleft\arraybackslash}X |
  | >{\raggedleft\arraybackslash}X |
  | >{\raggedleft\arraybackslash}X |
  | >{\raggedleft\arraybackslash}X |
  | >{\raggedleft\arraybackslash}X |}
 \hline
 Method & 
Number of extracted words & keywords/words & nouns(in non keyword set)/words & verbs(in non keyword set)/words & adj+adverbs(in non keyword set)/words\\
 \hline
 LUHN & 
276 & 10.5 & 40 & 35.5 & 14 \\
\hline
 YAKE &
 276 & 10.5 & 38.4 & 35.5 & 15.6 \\
\hline
OUR METHOD &
276 & 31 & 52 & 8 & 9 \\
\hline
\end{tabularx}
\label{table_russian}
\end{table*}
}

\section{Evaluation}
\label{more}

\subsection{Precision and Recall}

{
Results obtained for {\it Anna Karenina} are confirmed for several other texts; cf.~Table~\ref{tab_gogo}. We extracted for each text 
the same number of $\approx 300$ potential keywords via 3 methods: our method (implemented via Eqs.~(\ref{40}, \ref{41}, \ref{ade})), LUHN and YAKE. The precise number of extracted words depends on the text. 

As seen from Table~\ref{tab_gogo}, for long texts (with the length roughly comparable with {\it Anna Karenina}) our method outperforms both LUHN and YAKE in terms of the precision, i.e. the relative number of extracted keywords which is defined as the number of keywords extracted via the given method (for each text) divided over the total number of words proposed by the method as potential keywords. The advantage of our method is also seen in terms of recall, which is the number of keywords extracted via the given method (for each text) divided over the full number of keywords announced by an annotator for the text. (The definitions of precision and recall are reminded in Appendix \ref{prerecall}; in particular, the above results were found via Eqs.~(\ref{punjab}).) Importantly, for YAKE and LUHN the values of recall are lower than $0.5$, while for our method they are sizably larger than $0.5$ meaning that our method extracted the majority of potential keywords; see Table~\ref{tab_gogo}. 
}

In this context, we distinguish between long and short texts; cf.~Table~\ref{tab_gogo}. For short texts, our method needs modifications that are described above. After these modifications, our method implemented via Eq.~(\ref{sharik}) produces for short texts nearly the same results as LUHN and YAKE; see Table~\ref{tab_gogo}. For short texts, we extracted via each method the same number of $\sim 100$ words. However, our method still has an important advantage, since it allows us to extract topical groups of short texts nearly in the same way as for long texts; see Table \ref{hd1} where we analyze topical groups of {\it The Heart of Dog} by M. Bulgakov. We emphasize that this feature is absent for LUHN and YAKE.  

\comment{
\begin{table} 
\caption{For the database of 100 theses, we present the F1 evaluation results; e.g. the second column means that in 42 cases out of 100, all three methods produced the same F1 score. The third column means that in 23 cases the highest F1 results were produced by our method and Luhn, {\it etc}. }
\begin{tabular}{|l||l|l|l|l|l|l|l|l|}
\hline
{\footnotesize Method}          & {\footnotesize All 3 methods} & {\footnotesize LUHN+YAKE} & {\footnotesize Our+YAKE} 
& {\footnotesize Our+LUHN} & {\footnotesize LUHN} & {\footnotesize YAKE} & {\footnotesize Our} & {\footnotesize None} \\
\hline
{\footnotesize Number} &     {\footnotesize 42}        &  {\footnotesize 23}       &  {\footnotesize 2}       &  {\footnotesize 2}       
& {\footnotesize 6}   & {\footnotesize 1}   & {\footnotesize 16}  & {\footnotesize 8}   \\
\hline
\end{tabular}
\label{tab_100}
\end{table} 
}

\subsection{Inter-anontator agreement}
\label{inter2}

The performance of any keyword extraction method is evaluated by annotators. First, annotators should be provided with guidelines on the extraction process; e.g. characters are keywords, pay more attention to nouns and less to verbs and adjectives {\it etc}. Second, two (or more) annotators are independently given the set of keywords extracted by our algorithm from the same set of texts, and they mark words that they consider as keywords. So each annotator will get at the end a list of keywords {\it versus} non-keyword. Annotators are influenced by various subjective factors: background, prior knowledge, taste {\it etc}. However, the situation will not be subjective if different annotators produce similar results. To quantify the agreement between annotators we employed Cohen's kappa $\kappa$; see Table~\ref{tab_inter}. This statistical measure assesses inter-annotator agreement when working on categorical data in linguistics, psychology, and information retrieval; see \citep{cook_kappa} for review. It accounts for chance agreement and provides a more robust evaluation of agreement than the simple percentage. Cohen's $\kappa$ reads
\BEA
\label{kappa1}
&& \kappa=\frac{p_o-p_e}{1-p_e},~~ -1\geq \kappa\geq 1,\\
&& p_o=p(A=\kk,B=\kk)+p(A=\nk,B=\nk),\\
&& p_e=p(A=\kk)p(B=\kk)+p(A=\nk)p(B=\nk),~~
\EEA
where $p(A=\kk,B=\kk)$ is the joint probability for annotators $A$ and $B$ to identify keyword, $p(A=\nk,B=\nk)$ is the same for non-keywords (denoted by $\nk$), $p(A=\kk)$ is the marginal probability {\it etc}. 
Hence, $p_o$ is the agreement probability, while $p_e$ is the probability to agree by chance. Now $\kappa\to p_o$ for $p_e\to 0$, while for $p_e\lesssim 1$ even a relatively small, but positive difference $p_o-p_e$ is sufficient for $\kappa\to 1$.

Numerical interpretation of $\kappa$ is as follows \citep{cook_kappa}. No agreement: $\kappa<0$. Slight agreement: $0.2>\kappa>0$. Fair agreement: $0.4>\kappa>0.2$. Moderate agreement: $0.6>\kappa>0.4$. Substantial agreement: $0.8>\kappa>0.6$. Following these steps we got an agreement, which is between moderate and substantial both for short and long texts; see Table~\ref{tab_inter}. The agreement is better for global keywords, as expected. 

\subsection{Comparison with RAKE}

The method we developed was compared with two unsupervised keyword extraction methods that are among the best, LUHN and YAKE. As an additional comparison, let us take a look at RAKE (Rapid Automatic Keyword Extraction), another unsupervised method \citep{rake}. Its standard implementation in \url{https://pypi.org/project/python-rake/} requires that punctuation signs and stop-words are conserved in the text. It returns keyphrases that we partitioned into separate words. Here are the first (i.e. the highest score) 50 words extracted by RAKE for {\it Anna Karenina} (English version): 
{\it  finesses, ces, toutes, par-dessus, passer, influence, mot, nihilist, le, moral, disons, plaisir, h\"ubsch, recht, auch,
ich, bris\'e, est, en, moule, le, monde, du, merveilles, sept, les, jusqu, tomber, va, poulet, ce, blague, une, est, \c{c}a, tout, devoir,
votre, oubliez, vous, et, ausrechnen, sich, l\"asst, das, terre-\'a-terre, excessivement, mais, amour, le, filez, vous}. 
These words are mostly French and German (not English), and they are certainly far from being keywords. 

\subsection{Comparison with KeyBERT}

In section \ref{related} we mentioned KeyBERT, an unsupervised method that uses BERT to convert the input text and potential keywords into high-dimensional vectors (embeddings) \cite{bert}. These embeddings capture the semantic meaning of the words and phrases. BERT is a transformer deep neural network with at least 110 million parameters, which looks for the context of words by considering the entire sentence in a bidirectional way. KeyBERT generates a list of candidate keywords and keyphrases from the input text. For each candidate keyword/keyphrase, KeyBERT calculates the similarity between the embedding of the candidate and the embedding of the entire input text. High similarity means a higher score for a potential keyword.

We applied KeyBERT to {\it Anna Karenina}. To apply it efficiently, we extracted two-word keyphrases (which could then be split into different keywords if necessary). Below we present the 100 highest-score results of these applications together with their scores. Only 9 single keywords were actually extracted over 200 words: {\it anna, karenina, marriage, annushka, madame, peasant, karenins, sviazhsky}; i.e. the performance is much lower than for our method. 
The fact of two-word key-phrases provides additional information, but this information is of a specific type: it 
associates {\it anna} with words such as {\it husband, wife, marriage, household, love, married, sincerely, woman, madame, fashionable, courteously, unpardonable, emotionalism, lady}, {\it etc}. This provides some information about what {\it anna} does, or is involved in. We recall from our extensive discussion in section \ref{anna} that {\it Anna Karenina} is certainly not solely about Anna Karenina, but presents an epochal representation of Russian life at the end of 19th century.

Hence, when evaluated over long texts, KeyBERT performed worse than our method in terms of precision, recall, and topic extraction. We exemplified these facts on {\it Anna Karenina}, but they hold for several other long texts we checked.  

{\small
('anna husband', 0.4899);
('anna karenina', 0.489);
('anna wife', 0.4878);
('wife anna', 0.4833);
('household anna', 0.4813);
('karenina anna', 0.4755);
('husband anna', 0.4676);
('karenina marriage', 0.4615);
('love anna', 0.4607);
('anna involuntarily', 0.4606);
('superfluous anna', 0.4595);
('karenina husband', 0.4585);
('anna sincerely', 0.4533);
('married anna', 0.4499);
('care anna', 0.4464);
('anna married', 0.4455);
('anna annushka', 0.4446);
('anna commonplace', 0.4443);
('anna unnaturalness', 0.4424);
('arrange anna', 0.4422);
('anna strangely', 0.4417);
('anna love', 0.4415);
('anna irritates', 0.4413);
('anna peasant', 0.4406);
('anna courteously', 0.4406);
('woman anna', 0.4404);
('anna unpardonably', 0.4403);
('anna distress', 0.4401);
('dear anna', 0.4399);
('anna wonderfully', 0.4395);
('anna unmistakably', 0.4387);
('anna fashionable', 0.4382);
('madame karenina', 0.438);
('anna dear', 0.4379);
('anna lovely', 0.4375);
('anna emotionalism', 0.437);
('anna woman', 0.4367);
('annushka anna', 0.4363);
('distinctly anna', 0.4359);
('intensely anna', 0.4353);
('living anna', 0.4348);
('anna fascinate', 0.4344);
('perceive anna', 0.4344);
('recognize anna', 0.4343);
('anna rarely', 0.4343);
('maid anna', 0.4342);
('remarkable anna', 0.434);
('dress anna', 0.4326);
('irritate anna', 0.4323);
('karenina leo', 0.4302);
('person anna', 0.43);
('anna lady', 0.429);
('anna sister', 0.4281);
('repeat anna', 0.4276);
('anna occupation', 0.4276);
('anna meant', 0.4274);
('unusual anna', 0.4271);
('expression anna', 0.4268);
('anna karenin', 0.4264);
('karenins household', 0.426);
('turn anna', 0.4258);
('anna write', 0.4255);
('peasant anna', 0.4253);
('sincerely anna', 0.4253);
('unbecoming anna', 0.425);
('anna madame', 0.4245);
('occupation anna', 0.4242);
('discern anna', 0.4242);
('anna kindly', 0.4239);
('anna care', 0.4229);
('anna mentally', 0.4227);
('anna constantly', 0.4223);
('altogether anna', 0.4222);
('word anna', 0.4209);
('couple anna', 0.4209);
('special anna', 0.4208);
('anna interpose', 0.4204);
('anna query', 0.4203);
('anna dreamily', 0.4202);
('uttered anna', 0.4202);
('delight anna', 0.4199);
('indicate anna', 0.4196);
('description anna', 0.4195);
('anna anna', 0.4194);
('anna sviazhsky', 0.4187);
('complain anna', 0.4184);
('exceptional anna', 0.4182);
('anna irritable', 0.4175);
('anna properly', 0.4174);
('meet anna', 0.417);
('anna indifferently', 0.4167);
('anna manner', 0.4165);
('anna resolutely', 0.4164);
('anna memorable', 0.4163);
('anna wholly', 0.4161);
('anna sisterinlaw', 0.4158);
('anna special', 0.4158);
('articulate anna', 0.4153);
('explain anna', 0.4152);
('anna recognize', 0.4151)

}

\subsection{Comparison with supervised methods}

We compare our results with two supervised methods: KEA \citep{kea} and WINGNUS \citep{wingnus}. 
These two methods were selected, because they are relatively new, their software is free, and their structure is well-documented in literature. The implementation of both cases were
taken from pke python library: \url{https://github.com/boudinfl/pke/tree/master?tab=readme-ov-file}. Both methods were trained on the semeval2010 dataset available from \url{https://aclanthology.org/S10-1004/}. This dataset amounts to 284 abstracts from scientific articles included in this dataset, which were annotated by both authors and independent annotators.

The performance of KEA and WINGNUS was checked on long texts from Table~\ref{tab_gogo}. Their performance is generally lower than for our method. Here are two examples framed in terms of precision. For {\it Anna Karenina} KEA and WINGNUS led to 12.3\% and 10\%, respectively. For {\it Master and Margarita} these numbers are 15.3\% and 8.3\%, respectively; cf.~Table~\ref{tab_gogo}.

\subsection{Comparison with topic modeling methods}
\label{topos}

Above we described how human annotators can deduce topical groups of {\it Anna Karenina} using keywords extracted by our method. The same task -- known as topic modeling -- is achieved algorithmically (i.e., without human intervention) by several known NLP methods \citep{scikit}. We focused on 3 known topic modeling methods (all of them were applied without supervising), and below we compare their efficiency with our results. Before proceeding, let us note that topic modeling is generally applied to an entire corpus of texts, not just a single one. However, there are reasons to believe it can also be applied to a sufficiently long text. 

Non-Negative Matrix Factorization (NNMF) is an unsupervised method of topic modeling that applies to a single text; see Appendix \ref{external}. Its implementation for {\it Anna Karenina} produced the following 10 potential topics (their number is a hyper-parameter of the model). 

T-1: {\it levin, vronsky, anna, kitty, alexey, alexandrovitch,
arkadyevitch, stepan, room, wife}.

T-2: {\it walked, beauty, maid, gentleman, people, frou, complete,
carriage, completely, coming}

T-3: {\it living, sviazhsky, meaning, mare, kitty, book, natural,
listening, friends, suppose}

T-4: {\it read, work, ivanovitch, coat, drove, agriculture, lack,
hearing, matters, living}

T-5: {\it serpuhovskoy, received, young, desire, pass, asleep, set,
action, clerk, stay}

T-6: {\it doctor, stahl, coming, today, passion, porter, silence,
movement, object, levin}

T-7: {\it tanya, remember, game, deal, live, mamma, walking, bare,
easy, hurriedly} 

T-8: {\it noticed, possibility, christian, dmitrievitch, feelings,
fall, forget, success, stopped, suffer} 

T-9: {\it early, covered, recognized, angrily, connection, expression,
figure, breathing, nice, friend}

T-10: {\it scythe, nobility, elections, minutes, promised, extreme,
afraid, decided, ordered, lifting}

Only one of them (T-1) was reliable and approximately coincided with the first topical group (proper names of major characters) discussed above; see Table~\ref{tab_groups}. This is not surprising, since this topical group contains the most frequent content words of the text. Other topics extracted by NNMF turn out to be meaningless, i.e. they do not correspond to any topical group of the text.

We applied two alternative topic-modeling methods: Truncated SVD and LDA (Latent Dirichlet Allocation); see Appendix \ref{external}. Truncated SVD applied to a single text. LDA was attempted both for a single text directly and after training on a group of $\approx 100$ long texts. Both methods produced similar results: several topics were discovered, but all those topics were closely related to each other. Eventually, this situation amounts to only one sensible topic: the proper names of major characters. Let us illustrate this situation on the main topic discovered by LDA for {\it Anna Karenina}:
{\it levin, vronsky, anna, arkadyevitch, alexey, kitty, hand, stepan, long, alexandrovitch}.
In variations of this topic, {\it hand} can be changed to {\it wife}, or {\it brother}, {\it etc}. Likewise, LDA extracted (effectively) a single topic for other long texts, e.g. {\it Master and Margarita}. This topic is based on the following words: {\it margarita, ivan, hand, procurator, began, asked, woland, pilate, master, koroviev}. Again, besides {\it began} and {\it asked}, these are the proper names of the main characters.

In sum, NNMF, LDA, and truncated LDA produced consistent results for the studied long texts: they extracted a single topic based on the proper names of the main characters. This cannot be considered as a productive result.

\subsection{Comparison with the standard database}

We created all of the above databases in order to evaluate the method we employed. The standard databases should also be used in such cases. Hence, we analyzed a database of 100 theses [see \url{https://github.com/LIAAD/KeywordExtractor-Datasets?tab=readme-ov-file#theses}]. The average length of texts from this database is roughly comparable with the length of short texts from Table~\ref{tab_gogo}. 
Each of these texts was annotated manually and got 10 keywords. We also implemented LUHN, YAKE, and our method, each extracting 10 keywords from every database text. 
The evaluation of each method was based on the standard F1 score, which is the harmonic mean of the precision and recall; see (\ref{pushtu}) in Appendix \ref{prerecall}. The evaluation results are as follows. For 8 database texts, all three methods produced zero F1 score.  For the remaining 92 texts, LUHN wins (i.e., its F1 score is largest among three) in 73 cases, YAKE wins in 68 cases, and our method is the winner in 62 cases from 92, i.e. the performance of all 3 methods is comparable. 
(The concrete values of the F1 score can be looked up from \url{https://github.com/LidaAleksanyan/keywords_extraction_data/tree/master})
This is an expected conclusion for two independent reasons. First, the average length of the texts from the database is comparable with that of short texts from Table~\ref{tab_gogo}, where we also see roughly equal performance of all three methods. Second, the extracted number of keywords is small (i.e. 10). Even for longer texts, we would not see a well-defined advantage of our method for such a small number of extracted keywords. Unfortunately, standard databases do not have sufficiently many extracted keywords per text, even when the text is complex and structured. 

\section{Keyword extraction and distribution of words over chapters}
\label{chapters}

\begin{table*}\caption{First column: 36 words from {\it Anna Karenina} that have the highest score of YAKE \citep{yake,yake!}. Keywords are indicated by the number of their group; see Table \ref{tab_groups}. Among 36 words there are 25 non-keywords. Keywords refer mostly to group \1.\\
Second column: 36 words of {\it Anna Karenina} extracted via looking at the distribution of words over chapters, i.e. at the largest value of Eq.~(\ref{9}). Only 2 words out of 36 are not keywords. Several keyword groups are represented. }
 \begin{tabularx}{1\textwidth} 
{ 
  | >{\raggedright\arraybackslash}X 
  | >{\raggedleft\arraybackslash}X | }
 \hline
 36 words having largest score of YAKE  &
 36 words having largest values of Eq.~(\ref{9}) \\
 \hline 
levin\1, anna\1, vronsky\1, alexey\1, kitty\1, stepan\1, 
hand, alexandrovitch\1, smile, thought, arkadyevitch\1, time, 
love, face, eye, felt, man, feel, 
talk, life, answer, day, wife, begin, 
long, knew, turn, child, sergey\1, husband, 
work, princess\8, room, ivanovitch\1, people, woman 
&
levin\1, alexey\1, alexandrovitch\1, varenka\2, vronsky\1, kitty\1, doctor\8, stepan\1, scythe\6, anna\1,
arkadyevitch\1, marsh\6, countess\8, katavasov\2, priest\9, darya\1, veslovsky\2, alexandrovna\1, seryozha\1,
mare\6, sviazhsky\2, mihailov\2, brother, dolly\1, grass\6, sergey\1, princess\8, mow\6, marshal\7, konstantin\2, ivanovitch\2, peasant\6, lidia\1, sick, petritsky\2 \\ 
\hline
\end{tabularx} 
\label{tab_h}
\end{table*}

Long texts are frequently divided into sufficiently many chapters. It is an interesting question whether this fact can be employed as an independent criterion for extracting keywords. To search for such criteria, let us introduce the following basic quantities. Given a word $w$ and chapters $c=1,..,N_{\rm chap}$ we define $m_w(c)\geq 0$ as the number of times $w$ appeared in chapter $c$. Likewise, let $V_w(s)$ be the number of chapters, where $w$ appeared $s\geq 0$ times; i.e. $\sum_{s\geq s_0}V_w(s)$ is the number of chapters, where $w$ appears at least $s_0$ times. We have 
\BEA
&& {\sum}_{c=1}^{N_{\rm chap}}m_w(c)=N_w,\\
&& {\sum}_{s\geq 0}sV_w(s)=N_w,
\EEA
where $N_w$ is the number of times $w$ appears in the text; cf.~Eq.~(\ref{ordinary}). Hence, 
when taking a random occurrence of word $w$, we shall see $w$ appearing in chapter $c$ with probability
$m_w(c)/N_w$. Likewise, $sV_w(s)/N_w$ is the probability $w$ will appear in a chapter, where $w$ is encountered $s$ times.

It appears that quantities deduced from $m_w(c)/N_w$ do not lead to useful predictions concerning keywords. In particular, this concerns the entropy $-\sum_{c=1}^{N_{\rm chap}}\frac{m_w(c)}{N_w}\ln \frac{m_w(c)}{N_w}$ and correlation function $\sum_{c_1,c_2=1}^{N_{\rm chap}} |c_1-c_2|m_w(c_1)m_w(c_2)$ together with some of its generalizations. In contrast, the following mean [cf.~Eq.~(\ref{3}) with Eq.~(\ref{9})]
\BEA
\label{9}
{\sum}_{s\geq 0}\frac{s^2V_w(s)}{N_w},
\EEA
related to $sV_w(s)/N_w$ predicts sufficiently many global keywords; see Table~\ref{tab_h}. Similar results are found upon using the entropy $-{\sum}_{s\geq 0}\frac{sV_w(s)}{N_w}\ln \frac{sV_w(s)}{N_w}$ instead of Eq.~(\ref{9}). This formula is calculated for each word and then words with the largest value of Eq.~(\ref{9}) are selected. For {\it Anna Karenina}, at least the first 35-36 words selected in this way are keywords. Minor exclusions are seen in Table~\ref{tab_h}, which also shows that this method is much better than YAKE both in quantity and quality of keyword extraction. The advantage of this chapter-based method is that it does not depend on random permutations. Hence, it will be easier in practical implementations. The drawbacks are seen above: it depends on the existence of sufficiently many chapters (hence it certainly does not apply to texts with a few or no chapters), and it addresses only some of the keywords. 

\comment{We applied $V_w(s)$ to extract keywords that reflect the meaning of a text. Note that $V_w(s)$ effectively appears in scientometry, where it is employed for estimating the impact of scientists: the word $w$, chapters of the text, and $V_w(s)$ can be mapped to (resp.) a scientist, papers he/she produced, and the number of citations each paper got \cite{index}. Using this analogy, one can define for a word $w$ its h-index $h_w$. Now $h_w$ is the largest number with the following feature: $w$ appears in $h_w$ chapters, $h_w$ or more times \cite{index}. A bigger $h_w$ means that $w$ appears more in a larger number of chapters. However, $h_w=1$ if $h_w$ appears in all chapters once, and $h_w=1$ if $w$ appears only one time. As we checked, when it comes to extracting keywords, $h_w$ is certainly less useful than Eq.~(\ref{9}). Using the analogy back, it may turn out that (\ref{9}) (and related measures) find scientometric applications that improve the performance of the h-index. }

\section{Discussion and conclusion}
\label{final}

We proposed a method for extracting keywords from a single text. The method employs a spatial structure in word distribution. Our unsupervised method extends previous proposals, applies to a single text (i.e. it does not need databases), and demonstrates two pertinent applications: extracting the main topics of the text and separating between local and global keywords. For long texts, our analysis confirms that such a separation is semantically meaningful; see \citep{textual,pedro} for related ideas about various types of keywords. The method was illustrated in several classic literature texts; see Table~\ref{tab_gogo}. In researching the performance of our method we relied on expert evaluation of keywords (that show from moderate to substantial inter-annotator agreement). As expected, the agreement is better for global keywords. An important aspect of our method is that allows to extract topics of text by looking at keywords. In particular, we focused on the analysis of topics of {\it Anna Karenina} and {\it Heart of Dog}. 

\comment{
These methods also apply to a single text, and they do not require a text database; i.e. they are not of TF-IDF type \citep{review4}. LUHN is one of the oldest methods for keyword extraction. It uses the known fact that keywrods are frequently found among content words of a single text that are neither very frequent, nor very rare \citep{luhn}. Not very frequent is decided by omitting stop words, while not very rare is decided by following the thread of Zipf's law; see section \ref{lu}. YAKE is a more recent and rather popular method \citep{yake,yake!}. Its performance cannot be described in simple terms, though (as we witnessed) its performance does not differ much from that of LUHN at least when extracting single keywords. The reason for this is partially due to the use of Zipf's law in our implementation of LUHN.   }

Both in terms of precision and recall, our method outperforms several existing methods for keyword extraction. These methods include LUHN \citep{luhn}, YAKE \citep{yake,yake!} (which outperforms graph-based methods), RAKE \citep{rake}, KEA \citep{kea}, WINGNUS \citep{wingnus}, and KeyBERT \citep{bert}. We mostly compare with YAKE and LUHN, because these are our closest competitors.  
For sufficiently long texts, the advantage of our method compared to LUHN and YAKE is both quantitative (since it extracts $2-3$ times more words than LUHN and YAKE), and qualitative, because LUHN and YAKE do not extract the topics, do not distinguish between local and global keywords, and do not have efficient ranking of keywords. For shorter text only one advantage persists: our method helps to extract topical groups; see Table~\ref{tab_gogo} and Table \ref{hd1}. We show that our method generally extracts more nouns and longer content words than YAKE and LUHN. There are correlations between these features and the features of being a keyword; \cite{pnas,review4} provide additional arguments along these lines. Our method is also language-independent, as we checked with several translations of the same text. It shares this advantage with LUHN and YAKE. 

We demonstrated that our method of identifying topical groups of texts (where human annotators employ keywords extracted by our method) produced significantly better results than algorithmic methods of topic modeling such as Non-Negative Matrix Factorization, Truncated SVD, and LDA (Latent Dirichlet Allocation); see Appendix \ref{external}. 

We also worked out a method for keyword extraction that uses the fact that a text has sufficiently many chapters. This method is easy to implement and works better than LUHN and YAKE, but it is inferior to the previous one. However, we believe this method does have the potential for further development, e.g., for clarifying the conceptual meaning of keywords and relating it with higher-order textual structures. 

Our method is targeted to extract local and global keywords. Their spatial distribution moments have different behavior with respect to random permutations; see Eqs.~(\ref{40}--\ref{ade}). For short texts, the difference between local and global keywords is blurred; see Fig.~\ref{farmfig}. Put differently, local and global keywords are not so different from each other. Hence our method is not efficient (or more precisely nearly as efficient as YAKE and LUHN) in extracting keywords from short texts; see Table~\ref{tab_gogo} . We are currently searching for modifications of our method that can outperform those two methods also for shorter texts. We believe that this should be feasible because our method was able to extract topical groups of a short text; see Table \ref{hd1}.

\comment{
In this context, we are going to  modify the spatial mean and variance of the word [see Eqs.~(\ref{durnovo}, \ref{37})] such that they reflect the local frequency of the word. }

Future work will include adding n-gram analysis functionality to extract not only single words but also phrases of two or more words from a text. In particular, these key phrases are important for reflecting the aspects that were not studied in this work, i.e. relations of keywords to the style of the text. \cite{pagerank,papa,slava} provide useful information on how to find keyphrases from keyword extractions. Another useful idea for extracting keyphrases is to study co-occurrences between candidate keywords; see e.g. \citep{matsuo} for this technique. A more remote but important application will be to employ keywords for facilitating text compression methods \citep{allahverdyan2023optimal}. 

The present work leaves open the issue of finding an adequate model for random text, so that the implementation of a random permutation (on which our method relies) is not needed; see section \ref{kkk}. 
We were not able to employ theoretical models of a random text proposed by \cite{yngve,herdan,spatial1,spatial3}, because these models frequently imply asymptotic limits that are not reached for keywords. Further ideas about modeling the random gap distribution in non-asymptotic situations were proposed by \cite{zornig1,zornig2,pedro,zipf_pre}. We plan to address them in the future. Ideally, once the proper random model for the gap distribution is understood, random permutations will not be needed.

\section*{Acknowledgements}

This work was supported by SCS of Armenia, grant No. 21AG-1C038. It is a pleasure to thank Narek Martirosyan for participating in initial stages of this work. We acknowledge discussions with A. Khachatryan, K. Avetisyan and Ts. Ghukasyan. We thank S. Tamazyan for helping us with French texts. 

\bibliographystyle{ieeetr}
\bibliography{alpha}

\appendix

\section{Software employed}
\label{external}

-- Our data for texts from Table~\ref{tab_gogo} is freely available at \url{https://github.com/LidaAleksanyan/keywords_extraction_data/tree/master}. Our codes are available at \url{https://github.com/LidaAleksanyan/spatial_keyword_extraction}.

-- Natural Language Toolkit (NLTK) is available at \url{https://github.com/nltk/nltk}.

-- We used the following Python package with YAKE algorithm implementation:
\url{https://pypi.org/project/yake/}.

-- We used the following Python package for KeyBERT:
\url{https://pypi.org/project/keybert/}

-- For French and Russian texts we used (resp.) French LEFFF Lemmatizer 
\url{https://github.com/ClaudeCoulombe/FrenchLefffLemmatizer}, and
A Python wrapper of the Yandex Mystem 3.1 morphological analyzer pymystem3
\url{https://github.com/nlpub/pymystem3}.

-- Non-Negative Matrix Factorization (NNMF) was employed via 
\url{https://scikit-learn.org/stable/modules/generated/sklearn.decomposition.NMF.html}.

-- LDA (Latent Dirichlet Allocation) was employed via 
\url{https://radimrehurek.com/gensim/models/ldamodel.html}.

-- Truncated SVD was employed via \url{https://scikit-learn.org/stable/modules/generated/sklearn.decomposition.TruncatedSVD.html}.

-- The software for random permutations was taken from numpy.random.permutation:
\url{https://numpy.org/doc/stable/reference/random/generated/numpy.random.permutation.html}.

-- The implementations of KEA and WINGNUS were taken from \url{https://github.com/boudinfl/pke/tree/master?tab=readme-ov-file}. 

-- The semeval2010 dataset is available from \url{https://aclanthology.org/S10-1004/}. 

\section{Numerical illustrations of Eq.~(\ref{3})}
\label{cuba}

Here we illustrate on simple numerical examples how $C_2[w]$ defined via Eq.~(\ref{3}) can both increase or decrease after the distribution of words is made more homogeneous. We also illustrate that $C_1[w]$ [cf.~Eq.~(\ref{durnovo})] does not distinguish between homogeneous and inhomogeneous distributions. 
Take in Eq.~(\ref{3}) the following parameters: $N=10$ and $\ell=4$. Consider the following three distributions of words and the respective values of $C_2[w]$ ($\bar w$ means any word different from $w$):
\BEA
\label{ww1}
&& w\,\bar{w}\,\bar{w}\,\bar{w}\,\bar{w}\,\bar{w}\,\bar{w}\, w\, w\, w,\qquad C_2[w]=17, \qquad C_1[w]=3,\\
\label{ww2}
&& \bar{w}\,\bar{w}\,\bar{w}\,\bar{w}\,\bar{w}\,\bar{w}\,{w}\, w\, w\, w,\qquad C_2[w]=1,\qquad ~C_1[w]=1,\\
\label{ww3}
&& w\,\bar{w}\,\bar{w}\,{w}\,\bar{w}\,\bar{w}\,{w}\, \bar{w}\, \bar{w}\, w,\qquad C_2[w]=9, \qquad ~C_1[w]=3.
\EEA
Eq.~(\ref{ww1}) shows an inhomogeneous distribution of $w$, where $w$ appears both at the beginning of the text and its end. It has a large value of $C_2$. Eq.~(\ref{ww2}) is a strongly clustered distribution of $w$ with the minimal value of $C_2$. 
Eq.~(\ref{ww3}) shows the homogeneous distribution of $w$, its value of $C_2$ is intermediate between the above two. 
It is seen that $C_1$ does not distinguish between (\ref{ww1}) and (\ref{ww3}).

\section{Precision versus recall}
\label{prerecall}

For short texts, where the number of extracted and compared keywords is around 10-20, one employs the following standard definitions of precision and recall. For a given a text $T$, an annotator $A$ extracts set of keywords $w(T,A)$. Next, an automated method $M$ extracts a set $w(T,M)$ of candidate keywords. Let $N[w]$ be the number of elements in set $w$. Now define precision (${\rm Pre}$) and recall (${\rm Rec}$) for the present case as:
\BEA
{\rm Pre}(T,M,A)=\frac{N[w(T,A)\cap w(T,M)]} {N[w(T,M)]}, \qquad
{\rm Rec}(T,M,A)=\frac{N[w(T,A)\cap w(T,M)]} {N[w(T,A)]}.
\label{pushtu}
\EEA
Note that these definitions come from more general concepts, where the precision is defined as (relevant retrieved instances)/(all retrieved instances), while the recall is (relevant retrieved instances)/(all relevant instances). In application of these more general concepts to
(\ref{pushtu}) it is assumed that all words extracted by human annotator are by definition keywords.

However, definitions in (\ref{pushtu}) need to be modified for longer texts, where the automated methods offer many (up to 100-200) keywords, possibly joined in topical groups. We do not expect that human annotators (even those knowing the text beforehand) can extract {\it apriori} such a large amount of keywords more or less precisely. Hence, for long texts the experiments with annotators were carried out differently. Annotator $A$ studied all words extracted for a given text $T$ by all the involved $S$ methods $\{M_k\}_{k=1}^S$, and only after that completed his/her final list of words $\bar w(T,A)$. This set is larger (or equal) than the union of all keywords:
\BEA
\cup_{k=1}^S \bar w(T,M_k,A)\subset \bar w(T,A) ,
\label{khabar}
\EEA
where $\bar w(T,M_k,A)$ is the set of keywords identified by $A$ in a list of potential keywords extracted from $T$ using method $M_k$, and where $\bar w(T,A)$ can contain keywords that are not in $\cup_{k=1}^S w(T,M_k,A)$, though it can happen that $w(T,A)=\cup_{k=1}^S \bar w(T,M_k,A)$. 

Now the analogues of (\ref{pushtu}) are defined (for a given annotator $A$, and a given precision method $M$) as
\BEA
\overline{{\rm Pre}}(T,M,A)=\frac{N[\bar w(T,M,A) ]} {N[w(T,M)]}, \qquad
\overline{{\rm Rec}}(T,M,A)=\frac{N[\bar w(T,M,A)]} {N[\bar w(T,A)]}.
\label{punjab}
\EEA

Finally, remind that once recall and precision change between 0 and 1, a balanced way to account for both is to calculate their harmonic mean, which is standardly known as F1 score
\BEA
F1=2\,\frac{\overline{{\rm Pre}}(T,M,A)\overline{{\rm Rec}}(T,M,A)}{\overline{{\rm Pre}}(T,M,A)+\overline{{\rm Rec}}(T,M,A)}.
\label{gloria}
\EEA

\end{document}